\documentclass[12pt]{iopart}
\usepackage[numbers,sort&compress]{natbib}
\usepackage{lineno}
\usepackage{graphicx}
\usepackage[labelfont=bf,font=scriptsize]{caption}
\usepackage{hyperref}
\usepackage{booktabs}  
\usepackage{changepage,threeparttable} 
\usepackage{enumitem}
\usepackage{amsmath,amsthm}
\usepackage{multirow}
\usepackage[T1]{fontenc}

\usepackage{nicefrac}

\usepackage{thmtools}
\usepackage{thm-restate}
\newtheorem*{defn}{Definition}

\newcommand\bnfrac[2]{%
\scalebox{-1}[1]{\nicefrac{\scalebox{-1}[1]{$#1$}}{\scalebox{-1}[1]{$#2$}}}%
}
\usepackage[normalem]{ulem}
\usepackage{orcidlink}

\usepackage{iopams}
\usepackage{siunitx}
\makeatletter
\long\def\@makefntext#1{\parindent 1em\noindent 
 \makebox[1em][l]{\footnotesize\rm$\m@th{^\arabic{footnote}}$}%
 \footnotesize\rm #1}
\def\@makefnmark{\hbox{${^\arabic{footnote}}\m@th$}}
\def\@thefnmark{\arabic{footnote}}
\makeatother

\begin{document}
\title[Automation of quantum dot measurement analysis]{Automation of quantum dot measurement analysis via explainable machine learning}

\author{
Daniel Schug\textsuperscript{1}\orcidlink{0009-0001-3758-501X}, 
Tyler J. Kovach\textsuperscript{2}\orcidlink{0009-0007-0807-7300},
M. A. Wolfe\textsuperscript{2},
Jared Benson\textsuperscript{2}\orcidlink{0009-0009-1673-5259},
Sanghyeok Park \textsuperscript{2},
J. P. Dodson\textsuperscript{2}\orcidlink{0000-0003-4265-5024},
J. Corrigan\textsuperscript{2}, 
M. A. Eriksson\textsuperscript{2}\orcidlink{0000-0002-3130-9735},
Justyna P. Zwolak\textsuperscript{3,4,5*}\orcidlink{0000-0002-2286-3208}
}

\address{\textsuperscript{1}Department of Chemistry and Biochemistry, University of Maryland, College Park, Maryland 20742, USA}
\address{\textsuperscript{2}Department of Physics, University of Wisconsin-Madison, Madison, Wisconsin 53706, USA}
\address{\textsuperscript{3}National Institute of Standards and Technology, Gaithersburg, MD 20899, USA}
\address{\textsuperscript{4}Joint Center for Quantum Information and Computer Science, University of Maryland, College Park, Maryland 20742, USA}
\address{\textsuperscript{5}Department of Physics, University of Maryland, College Park, Maryland 20742, USA}

\ead{jpzwolak@nist.gov}

\begin{abstract}
The rapid development of quantum dot (QD) devices for quantum computing has necessitated more efficient and automated methods for device characterization and tuning.
Many of the measurements acquired during the tuning process come in the form of images that need to be properly analyzed to guide the subsequent tuning steps. 
By design, features present in such images capture certain behaviors or states of the measured QD devices.
When considered carefully, such features can aid the control and calibration of QD devices.
An important example of such images are so-called \textit{triangle plots}, which visually represent current flow and reveal characteristics important for QD device calibration.
While image-based classification tools, such as convolutional neural networks (CNNs), can be used to verify whether a given measurement is \textit{good} and thus warrants the initiation of the next phase of tuning, they do not provide any insights into how the device should be adjusted in the case of \textit{bad} images. 
This is because CNNs sacrifice prediction and model intelligibility for high accuracy.
To ameliorate this trade-off, a recent study introduced an image vectorization approach that relies on the Gabor wavelet transform (Schug \textit{et al.} 2024 \textit{Proc. XAI4Sci: Explainable Machine Learning for Sciences Workshop (AAAI 2024) (Vancouver, Canada)} pp 1–6). 
Here we propose an alternative vectorization method that involves mathematical modeling of synthetic triangles to mimic the experimental data. 
Using explainable boosting machines, we show that this new method offers superior explainability of model prediction without sacrificing accuracy. 
This work demonstrates the feasibility and advantages of applying explainable machine learning techniques to the analysis of quantum dot measurements, paving the way for further advances in automated and transparent QD device tuning.
\end{abstract}
\vspace{2pc}
\section{Introduction}
\label{sec:intro}
In many machine learning (ML) applications, there has been a longstanding trade-off between the accuracy and interoperability of candidate models~\cite{Luo19-BAI, Zhang19-TRA, Baryannis19-SPI}. 
This is evident in the extreme example of deep neural networks (DNNs), which can offer excellent accuracy for many problems, often surpassing existing methods~\cite{Krizhevsky12-CNN, He15-DSH, He16-DRL}.
Yet, the best-performing ML models are limited in their interpretability due to the number of inaccessible layers.
Alternatively, simple techniques like linear models or decision trees allow the user to fully comprehend the internal weights.
However, these techniques often cannot model the complex relationships seen in modern datasets. 
For image data, there has been considerable progress toward finding a middle ground, typically through explaining complex models with surrogates such as the Local Interpretable Model-agnostic Explanations (LIME)~\cite{lime} and Shapley~\cite{NIPS2017_7062}. 

In contrast with many black-box ML models, explainable boosting machines (EBMs) are a glass-box method that enables the model to be directly interpretable rather than relying on surrogate explanations \cite{lou2013accurate}. 
Specifically, EBMs extend generalized additive models to include pairwise interactions, allowing one to observe the relationship between features. 
EBMs often provide accuracy on par with many black-box models with the additional advantage of enhanced intelligibility, which makes them an appealing replacement for other models, especially in applications of consequence, such as medicine~\cite{Caruana15-IMH, Letham15-ICB, Ustun15-SLI} or finance~\cite{Israel20-CML}. 
However, to date, EBMs have not been adapted to any other data type than tabular data.

For spatial data, such as images, interpretability is more challenging.
This is partly due to their composition by structures highly correlated at multiple scales in the two-dimensional (2D) space. 
This feature is one of the main reasons why convolutional neural networks (CNNs)~\cite{Krizhevsky12-CNN} and, more recently, vision transformers (ViTs)~\cite{Dosovitskiy21-TIR} have quickly become the dominant ML approach for many computer vision tasks.
However, the black-box nature of CNNs and ViTs makes their use prohibitive in applications where a solid understanding of model predictions is necessary.
While making CNNs and ViTs more interpretable has been a very active area of research, techniques proposed to date vary in utility in many applications, especially as the depth of the neural network increases \cite{Zhang18-VID, Kim22-IVT, Pan21-IVT}. 

Our previous work developed a methodology that addresses some of these interpretability concerns for measurement data by combining image vectorization with EBMs.
Using EBMs as models for image data poses numerous challenges, the principal of which is the mapping from images to a vector representation that could be used directly with EBMs. 
To achieve this goal, we used the Gabor Wavelet transform and a constrained optimization procedure to extract key image features from the data.
We also applied custom feature engineering to tailor this process to the particular dataset~\cite{Schug24-ECT}. 
To ensure that the resulting model produces human-agreeable interpretations, we relied on domain knowledge and understanding of the physical systems under investigation to inform the feature extraction process. 
Here, we demonstrate that the same approach can be successfully applied to assist in the tune-up of accumulation mode Si/Si$_{x}$Ge$_{1-x}$ quantum dot (QD) devices. 
We also propose an alternative image vectorization method involving the generation of synthetic data to approximate the experimentally acquired scans~\cite{Schug23-EES}.
We then show that both methods result in comparable performance, but the latter produces more intuitive and easier to interpret features.

The paper is organized as follows: in section~\ref{sec:background}, we provide a brief overview of the scientific context of the problem.
Section~\ref{ssec:qd_tuning_background} describes the problem of tuning QD devices and introduces the concept of triangle plots. 
Data used for benchmarking is discussed in section~\ref{ssec:tr_plot_data}.
The two vectorization methods, the Gabor filterbank approach and the synthetic data modeling approach, are described in section~\ref{ssec:gf_approach} and section~\ref{ssec:stf_approach}, respectively.
Finally, the EBMs performance on experimental data vectorized using both methods, as well as using a hybrid approach where the dominant Gabor filter is combined with synthetic data, is presented in section~\ref{sec:results}.
We conclude with a discussion of the future direction in section~\ref{sec:conclusion}.

\section{Background and methods}
\label{sec:background}
In this section, we introduce the scientific context of our problem and the data utilized in our experiments.
We also present an overview of the signal processing methods used in the \textit{Gabor filterbank approach} and the \textit{synthetic data modeling approach}.

\subsection{Quantum dot tuning problem and triangle plots}
\label{ssec:qd_tuning_background}
\begin{figure}[t]
  \centering
  \includegraphics[width=0.95\linewidth]{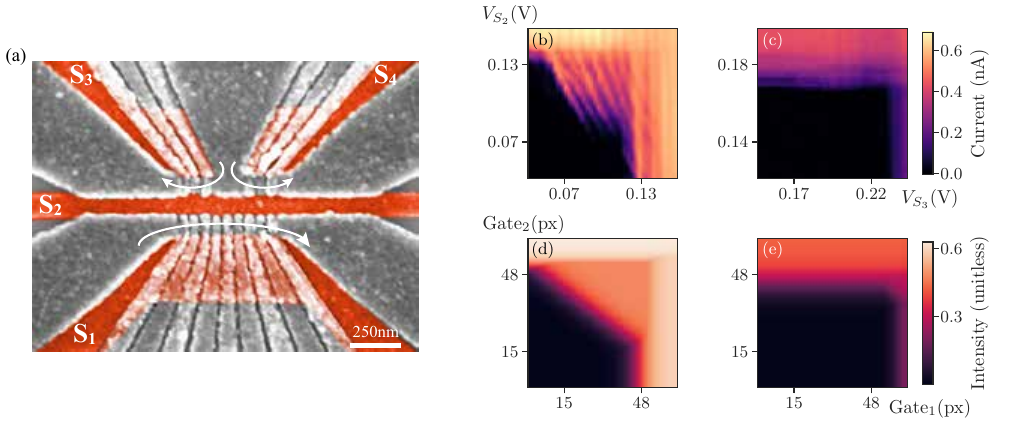}
  \caption{(a) False-colored SEM micrograph of a typical Si/Si$_{x}$Ge$_{1-x}$ heterostructure device. 
  The red highlighted gates are the screening gates that are swept for the triangle plot. 
  This quad-QD device has three distinct current channels between the screening gates marked with white arrows. 
  One of the channels (in the lower half of the device) contains four QDs in series while the other two channels (in the upper half of the device) are single QD channels for charge-based single-electron-transistor readout.
  (b) An example of a good triangle plot with high visibility in the triangle region above the background and (c) a corresponding synthetic triangle plot. 
  (c) An example of a bad triangle plot with little to no current in the triangle region and (d) a corresponding synthetic triangle plot.}
  \label{fig:dev_data}
\end{figure}

Arrays of QDs --  interconnected islands of electrons confined in a semiconductor heterostructure with unique properties that allow them to act as artificial atoms -- are a leading candidate for use as qubits, the fundamental information carriers in quantum computers~\cite{Burkard21-SSQ}. 
We use a type of gate-defined QD device, as shown in figure~\ref{fig:dev_data}(a), in which three layers of overlapping gates fabricated on top of a Si/Si$_{x}$Ge$_{1-x}$ heterostructure are used to form and control QDs~\cite{Zajac16-SGA, Dodson:2020p505001}.
Scaling these systems to large arrays suitable for quantum computations is a challenging task, as with the number of QDs, the number of gates needed to control them grows, making the manual tuning process unfeasible. 
An autotuning framework incorporating ML tools was originally proposed and validated off-line in Ref.~\cite{Kalantre17-MLD} using premeasured experimental scans capturing a large range of gate voltages and then deployed online (i.e., \emph{in situ}) to tune a double QD in real-time in Ref.~\cite{Zwolak20-AQD}.
A detailed description of the tuning process is available in Ref.~\cite{Zwolak21-AAQ}.

One sub-problem within the tuning procedure is determining the voltage placement of gates to allow the formation of isolated current channels inside the 2D electron gas formed at the intersection of the Si and Si$_{x}$Ge$_{1-x}$ layers in the heterostructure, see figure~\ref{fig:dev_data}(a). 
One way to achieve proper gate voltages is by sweeping certain gates, which we call screening gates, [S$_1$, S$_2$, S$_3$, and S$_4$ in figure~\ref{fig:dev_data}(a)] until the currents begin flowing in the correct channels. 
The bright bars at the edges of the images in figure~\ref{fig:dev_data}(b) and (c), called \textit{walls}, are due to the current flowing strictly under those gates.
The region with the ridged pattern figure~\ref{fig:dev_data}(b) is the area of interest -- it indicates that the current is not flowing under any screening gates, but instead, it is flowing between them. 
If there is no current in this region, as shown in figure~\ref{fig:dev_data}(c), it is necessary to increase the voltage of gates over the current channel to accumulate more electrons in the 2D electron gas and try again.

In gate-defined QD devices, the area of interest usually takes the form of a triangle -- thus the name \textit{triangle plots} -- and indicates the formation of an isolated current channel. 
Due to the physics principles at play, the only possible orientation of the triangle plot is the one with the right angle in the top-right corner of the image.
The patterned texture inside this region is due to charge defects and variability in gate uniformity near the current channel~\cite{Ye24-CCF}. 
Combining this with the fact that there is no complete analytical form for the triangle plot at this time makes physical modeling of the measurement impossible.
While researchers can easily visually identify these regions, finding them automatically using simple threshold analysis is challenging; largely due to experimental variations for different voltage configurations and when measuring different pairs of gates.
In moving towards automating the tuning process, it is desirable to have an algorithm that can (1) predict whether the triangle plot indicates a well-behaving current channel and (2) explain this prediction.

\subsection{Triangle plots dataset}
\label{ssec:tr_plot_data}
The images in the triangle plots dataset represent various scales and orientations of image data as well as variations in the triangle region size.
About $90~\%$ of the images were taken at the high temperature of $1.3$~\si{\kelvin}, instead of roughly $100$~\si{\milli\kelvin}, which means that the Coulomb blockade and other gate turn-ons are much less steep than in typical measurements. 
Since the high-intensity features (e.g., the Coulomb blockade fringes), while prominent to the eye, are less predictive of correct device performance, we focus on the larger-scale features, such as the presence or absence of the triangle region.

For our study, images that contain the triangle region, as shown in figure~\ref{fig:dev_data}(a), are considered \textit{good}.
Images that do not contain the triangle region are considered \textit{bad}. 
We use 902 experimentally acquired triangle plots, of which a total of 210 are labeled as having good triangles and 692 are labeled as having bad triangles.

The images range in size from $31 \times 31$ pixels to $171 \times 151$ pixels.
Prior to analysis, all images are resized to $64 \times 64$ pixels, utilizing a bicubic interpolation provided by the Pillow image library~\cite{clark2015pillow}.
This is done to enable using fixed-parameter filterbanks and synthetic triangles with comparable parameter ranges.
While bicubic rescaling can distort the angle of the diagonal region, the majority of the samples that needed preprocessing were in the bad class and thus had minimal to no diagonal activity.
In the good class, the maximum observed aspect distortion is within about $13~\%$, which corresponds to about $3.5^\circ$ difference and does not present significant errors in the context of the Gabor filters.
Each image is then vectorized following the procedure described in the Gaber filterbank approach (section~\ref{ssec:gf_approach}) and the synthetic data modeling approach (section~\ref{ssec:stf_approach}).
The QFlow Triangles dataset containing all experimental images as well as the corresponding synthetic data and Gaber filterbanks is available at Zenodo~\cite{qf-triangles}.

\subsection{Gabor filterbank approach}
\label{ssec:gf_approach}
To vectorize the image data, we utilize the 2D Gabor wavelet transform, an oriented multi-scale representation that is shown in Ref.~\cite{daugman_1988} to be a model for complex neurons in mammalian vision. 
Gabor wavelet transform has seen frequent use in numerous computer vision tasks and serves as an economical and effective feature transform. 
As such, this representation seems ideal for extracting oriented features, as well as textures, from image data. 

\begin{defn}
    The 2D Gabor Kernel for parameters $p = \{\sigma_x,\sigma_y,\lambda,\theta\}$ for $(x,y) \in \mathbb{R}^2$ is defined as
    \begin{equation}
        G_{p}(x,y) = \frac{1}{\sqrt{2\pi}\sigma_x\sigma_y} \, {\rm e}^{-\frac{1}{2}(\frac{x^2}{\sigma_x^2} + \frac{y^2}{\sigma_y^2})} \, 
        {\rm e}^{i \lambda(x \sin(\theta) + y \cos(\theta))},
    \end{equation}
    where $\sigma_x$ and $\sigma_y$ represent scale in $x$ and $y$, and $\theta$ and $\lambda$ are the wave direction and wavelength.
    We further denote the convolutional application of kernel $G_p$ to an image $u(x,y)$ to be $G_p(x,y) * u(x,y)$.
\end{defn}
Specifically, we consider small filterbanks of Gabor wavelets to capture scales and orientations directly relevant to the triangle plots image data.
Relying on wavelets leads to a significantly more compact representation. 

\begin{defn}
        The 2D Gabor filterbank for discrete set of $N$ parameters $P = \{p_0,...,p_N\}$ 
    \begin{equation}
        G_P(x,y) = \{G_{p_i}(x,y)\}_{i=0}^N.
    \end{equation}
    We denote the application of the filterbank $G_P$ to an image $u$ to be $G_P * u = \{G_{p_i}(x,y) * u(x,y)\}_{i=0}^N $
\end{defn}

In practice, we attempt to construct the set of parameters $P$ such that the Fourier transform of the filterbank $G_P$ supports prominent frequencies observed in the Fourier transform of the image $u$.
This can be accomplished with optimization or other filterbank construction techniques~\cite{Daugman04-IRW, Kruger02-GWH, Kwolek05-FGF}.

To produce the vector representation of the image, we take advantage of heuristics.
In particular, the main factor allowing the discrimination of good and bad triangles is the presence of the pattern of line-like features at approximately $45^{\circ}$ orientation corresponding to the interaction between the two gates, see figure~\ref{fig:dev_data}(b). 
The image is vectorized by taking the $L^2$-norm of each filter in the Gabor filterbank with this orientation and at different scales. 
This achieves a single numerical measure of the filter's response to the given image, which is ultimately descriptive due to the images' simple structural content.
The specific scales used were obtained through an iterative process.
At each iteration, an EBM model is trained on an available filterbank, starting with one that consists of a large number of possible scales and is refined by eliminating scales that are not discriminative.
The process is repeated until no further reduction of the filterbank size is observed. 
The resulting filterbank contains scales determined to be highly discriminative. 

In our case, the final filterbank consists of six filters at scales $\sigma_{x}=\sigma_y=4, 8, 16$, and orientation $45^{\circ}$ and $-45^{\circ}$ degrees, where $\lambda$ is fixed to $1$. 
We also perform additional feature engineering in the form of extracting the estimated location of narrow, edge-like Gabor filters to localize the boundaries of the walls and triangle regions. 
This technique is used to absorb important spatial features for later use in classification. 
While other techniques for vectorizing the data based on orientation or scale are possible and, in some instances, advantageous, we chose Gabor because it enables encapsulating the specific diagonal behavior in very few filters. 
Such vectorization is desirable to ensure that the prediction interpretation is given in terms of relevant features.

\begin{figure}[t]
  \centering
  \includegraphics[width=0.95\linewidth]{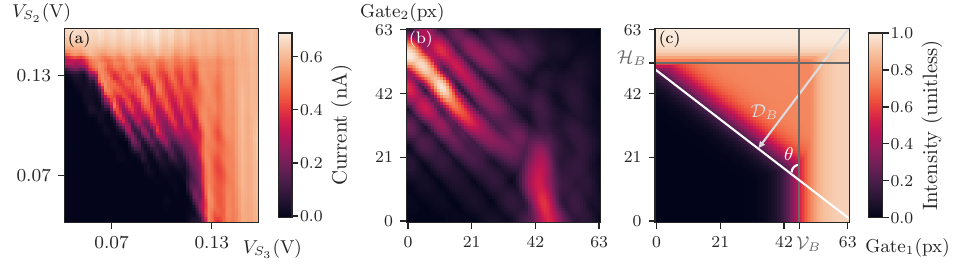}
  \caption{(a) A sample experimentally acquired scan, the same as in figure~1(b).
  (b) The Gabor filters for the scan shown in (a), weighted by their importance as indicated by the EBM model.
  (c) Synthetic fit to scan shown in (a).
  The $\mathcal{H}_B$, $\mathcal{V}_B$, and $\mathcal{D}_B$ are the horizontal, vertical, and diagonal optimal boundaries, respectively, and the $\theta$ indicates the orientation of the diagonal boundary.}
  \label{fig:methods_vis}
\end{figure}

Figure~\ref{fig:methods_vis}(b) shows a weighted sum of the filters of the final filterbank for a scan shown in figure~\ref{fig:methods_vis}(a), with weights defined by the contribution of each term in the EBM. 
While the resulting filterbank is an incomplete representation of the data, it covers primary discriminant support in the frequency domain between classes. 
The features are designed to measure the extent of response of the diagonal component of the triangle plot at different scales, where we operate under the intuition that good triangles will have a greater associated response at the $\pm45^{\circ}$ orientations than bad triangles, within a region resembling a fringed isosceles right triangle. 
The location of the edge-like filters acts as a further sieve for the presence and quality of a triangle region. 
While other vectorization techniques might provide more complete representations of the data and, in some cases, offer superior classification performance, this effect can be diminishing while harming the overall interpretability.

\subsection{Synthetic triangle plots} 
\label{ssec:stf_method}
To enable the creation of interpretable features, we generate a set of crude synthetic triangle plots.
The synthetic data share certain salient features visible in experimental triangle plots, namely the presence of walls of varying width and a possible diagonal region. 
It is important to note that while the resulting images are visually similar to experimental data, as depicted in figure~\ref{fig:dev_data}(c) and (e), this approach does not produce physically realizable data.

For generating the synthetic triangles, we utilize the 2D sigmoid function,
\begin{equation}
    s_{\rm m,r,b}(x) = \frac{\rm m}{1+{\rm e}^{{\rm r}(x-{\rm b})}},
\end{equation}
where ${\rm m}$, ${\rm r}$, and ${\rm b}$ control the magnitude, rate, and shift of the sigmoid, respectively.
The motivation for this choice is the resemblance of the idealized walls and triangle region to the step function.
In practice, the transition from no signal to wall and triangle should be smooth.
Triangle plots are assumed to have a horizontal and/or vertical wall corresponding to individual gates, as well as a possible third diagonal region linking the walls.
To construct the 2D triangle plots, we compose three sigmoids:
\begin{equation}
    \Delta_{(x,y)} = \max\big(s_{\rm h}(x), s_{\rm v}(y),
    s_{\rm d}(x\sin\theta + y\cos\theta)\big)
\end{equation}
where $s_{\rm h}$, $s_{\rm v}$, and $s_{\rm d}$ are the parameters for the horizontal, vertical, and diagonal sigmoid, respectively, with $\theta$ indicating the orientation of the diagonal sigmoid. 
In practice, we evaluate this function on an evenly spaced grid of $(x,y)$ points. 

It is important to restrict the parameters used to define the synthetic triangles to physically reasonable ranges.
The simple constraints we chose include a requirement that the sigmoids cannot be infinitely dilated and that the walls must be in the image and have positive heights.
The criteria used to determine the presence of a triangle region are derived from the definition of the triangle plot: we require that $s_{\rm d}(x\sin(\theta) + y\cos(\theta))$ is greater than $s_{\rm h}$ and $s_{\rm v}$.
We also impose a constraint that the orientation of the diagonal sigmoid is sufficiently diagonal, with the cutoff defined as $0 \leq \theta \leq \nicefrac{\pi}{2}$.

\subsection{Synthetic data modeling approach} 
\label{ssec:stf_approach}
The synthetic triangles are used to model the experimental triangle plots by deriving from experimental data parameters defining a most similar synthetic triangle.
The resulting vector provides a powerful and compact representation for ML.
Moreover, this vectorization approach has a key advantage in that the vector features correspond directly to the original image and that fewer assumptions about the frequency content of the data are made.

We start by defining an appropriate measure of image similarity. 
The experimentally acquired triangle plots contain considerable noise, texture, and other high-frequency features that are not essential in characterizing the overall structure. 
The synthetic triangles can serve as a representation of the boundaries of the vertical, horizontal, and diagonal walls defined by a shifted, scaled, and dilated sigmoid function. 
Since the synthetic and experimentally acquired plots are assumed to be structurally similar by design, both can be transformed into a region of the frequency domain that they are expected to share. 

In practice, this involves applying an ideal low-pass filter, such as a Gaussian filter. 
To focus on the details of the structure, we use the magnitude gradient of the low-pass filtered image. 
While this can be thought of as a simple edge detection, we do not have \textit{a priori} knowledge about the appropriate scale of the Gaussian for any particular image. 
The presence of the low-pass filter makes this approach notably more insensitive to certain classes of noise, which is desirable for our use case.
To quantify the similarity between an experimentally acquired and synthetic triangle plot, we use the $L^2$-norm applied to both transformed images, assuming that the two images are visually similar if their distance in this transformed space is small. 

\begin{defn}
    Let $U$ and $V$ be two images subject to some transform $F$ with parameters $q$.
    The similarity measure between $U$ and $V$ is defined as 
    \begin{equation}
        \mathcal{S}_q^F(U,V) = ||F_q(U)-F_q(V)||_2
    \end{equation}
\end{defn}

The desired vector representation of an experimentally acquired image $\mathcal{I}_{\rm exp}$ is obtained by optimizing the parameters defining the corresponding synthetic image $\mathcal{I}_{\rm sim}$ such that $\mathcal{S}_q^F(\mathcal{I}_{\rm exp}, \mathcal{I}_{\rm sim})$ is minimized subject to boundary constraints.
We set up the optimization problem as:
\begin{equation}\label{eq:optim}
    q^*,\sigma^* = \arg\!\min_{q,\sigma} \big\{ \lambda\,\mathcal{S}_{q,\sigma}^{\nabla_\Gamma}(\mathcal{I}_{\rm exp}, \mathcal{I}_{\rm sim})
    + (1-\lambda) \, \mathcal{S}_{q}^{I}(\mathcal{I}_{\rm exp}, \mathcal{I}_{\rm sim}) + \epsilon(q) \big\},
\end{equation}
where $\mathcal{S}_\sigma^{\nabla_\Gamma}$ is the similarity measure using the gradient of the Gaussian transform, $\nabla_\Gamma(U)=\nabla(\Gamma_\sigma(U))$; $\mathcal{S}^{I}$ is the similarity measure using the identity transform, $I(U)=U$; $\lambda$ is a hyperparameter giving us control over the balance between trusting $\mathcal{S}_\sigma^{\nabla_\Gamma}$ and $\mathcal{S}^{I}$ similarity measures; and
\begin{equation}
    \epsilon(q) = 
    \begin{cases}
        C, \quad \text{if $q$ violates boundary constraints,}
        \\
        0, \quad \text{otherwise,}
    \end{cases}
\end{equation}
for some arbitrarily large penalty constant $C$.
In practice, it is desirable to make this penalty larger than the other terms of the objective function so that the constraints are satisfied.

Since the initial guess for optimization is determined solely based on the raw image, we use the differential evolution global optimizer implemented in SciPy~\cite{scipy} to remedy this limitation. 
This makes our method less sensitive to initial values than if we had used a local optimizer but incurs a greater computational cost. 
A local optimizer might still be viable in cases where the phenomenological model is simpler, e.g., with fewer parameters or with guaranteed convexity.
In an experimental context, we have access to additional data tied to each image to aid in establishing initial guesses for optimization.

The optimal parameters, as well as the value of the objective function using those parameters, define the final vector representation of image $\mathcal{I}_{exp}$:
\begin{equation}
    \textbf{v}(\mathcal{I}_{exp})=[\sigma^*, \mathcal{H}_{(B,M,R)}(q^*), \mathcal{V}_{(B,M,R)}(q^*), \mathcal{D}_{(B,M,R,\theta)}(q^*), \mathcal{F}(q^*)]
\end{equation}
where $B$, $M$, and $R$ denote the boundary, magnitude, and rate for the optimal horizontal $\mathcal{H}$, vertical $\mathcal{V}$, and diagonal $\mathcal{D}$ component, respectively; $\theta$ indicates the orientation of the diagonal boundary; $\sigma^*$ is the optimal scale, $q^*=[\mathcal{H}, \mathcal{V}, \mathcal{D}]$[see. Eq.~(\ref{eq:optim})]; and $\mathcal{F}$ is the value of the objective function for the optimized parameters (the fit fitness). 
For a visualization of these features, see figure~\ref{fig:methods_vis}(c).

\section{Results}
\label{sec:results}
We present the results of experiments with the Gabor filterbank approach, the synthetic data modeling approach, and the hybrid approach combining the synthetic features with the single most informative Gabor feature.
The transformation procedures described in sections~\ref{ssec:gf_approach} and~\ref{ssec:stf_approach} are applied to each experimentally observed triangle to produce the final vector of features $\textbf{v}(\mathcal{I}_{exp})$. 
In each case, the vectorized features $\textbf{v}(\mathcal{I}_{exp})$ are used to train and test an EBM model. 
Since the dataset is quite imbalanced in terms of good and bad data, the training is performed using five six-fold stratified cross-validations, with $10~\%$ of the data withheld for testing purposes.
To determine the optimal $k$-fold configuration, we performed the stratified cross-validation for $k$ ranging between $1$ and $10$, as we found that $k=6$ achieved the most performant set of models.

\begin{table}[t]\flushright
\scriptsize
\captionsetup{width=\linewidth,font=footnotesize}
\caption{The results of five six-fold stratified cross-validation for the experimentally acquired data.
The accuracy, type I error, and type II error are reported for each method. 
Corresponding confusion matrices are included for completeness.
The value(uncertainty) notation is used to express uncertainties. 
All uncertainties herein reflect the uncorrelated combination of single-standard deviation statistical and systematic uncertainties.}
\label{tab:model_comparison}
\begin{tabular*}{\textwidth}{@{} p{2.8cm}@{\extracolsep{\fill}}cccc@{}} 
\toprule
\textbf{Model} & \textbf{Accuracy} [\%] & \textbf{Type I} [\%] & \textbf{Type II} [\%] & \textbf{Confusion matrix} [counts]\\ 
 \midrule
\textbf{Gabor} \textbf{filterbank} & 92.5(1.2) & 4.9 (0.9) & 2.6 (0.7) &
 \begin{tabular}{ c|c|c } 
$\bnfrac{\,\,\,{\rm pred}\rightarrow}{{\rm real}\downarrow\,}$ & Good & Bad \\ \cline{1-3}
 Good & {\textbf{406 (4)}} & 14 (4) \\ \cline{1-3}
 Bad & 27 (5) & {\textbf{99} (5)} \\ 
\end{tabular}
\\ 
 \midrule
\textbf{Synthetic} \textbf{triangles} & 90.5(1.0) & 5.8 (0.5) & 3.6 (0.6) &
 \begin{tabular}{ c|c|c } 
$\bnfrac{\,\,\,{\rm pred}\rightarrow}{{\rm real}\downarrow\,}$ & Good & Bad \\ \cline{1-3}
Good & {\textbf{400 (3)}} & 20 (3) \\ \cline{1-3}
Bad & 32 (3) & {\textbf{94} (3)} \\ 
\end{tabular}
\\
 \midrule
\textbf{Hybrid} \textbf{approach} & 91.7(8)  & 5.0 (0.8) & 3.3 (0.2) &
 \begin{tabular}{ c|c|c } 
$\bnfrac{\,\,\,{\rm pred}\rightarrow}{{\rm real}\downarrow\,}$ & Good & Bad \\ \cline{1-3}
Good & {\textbf{402 (1)}} & 18 (1) \\ \cline{1-3}
Bad & 27 (4) & {\textbf{4.3} (4)} \\ 
\end{tabular}
\\ 
 \bottomrule
\end{tabular*}
\end{table}

To improve the accuracy and interpretability at a modest training time cost, we utilize the smoothing and greedy rounds parameters~\cite{nori2019interpretml}. 
For a more thorough discussion of this, see Ref.~\cite{nori2021dp}.
We carry out five six-fold stratified cross-validations and report averaged results as well as type I and type II errors for each method.
For completeness, we also include the averaged confusion matrices.

The results from all experiments, presented in table~\ref{tab:model_comparison}, show 
a relatively close performance for all three methods, with the Gabor fitting approach only slightly outperforming the more intuitive synthetic data modeling approach, at $92.5(1.2)~\%$ vs. $90.5(1.0)~\%$, respectively.
Interestingly, the inclusion of a single, most important Gabor filter in the vector of features obtained using synthetic data modeling brings the EMBs performance back up to $91.7(8)~\%$. 
Importantly, the model resulting from the hybrid vectorization retains the majority of the performance while the features in the hybridized vector remain well aligned with the physical intuition of the problem.
This validates that the optimal fit captures the location and average intensity of the walls and diagonal components.
Modifying the diagonal component function to capture the ridged pattern observed in the experimental data could further improve the performance.

\begin{figure}[t]
  \centering
  \includegraphics[width=0.95\linewidth]{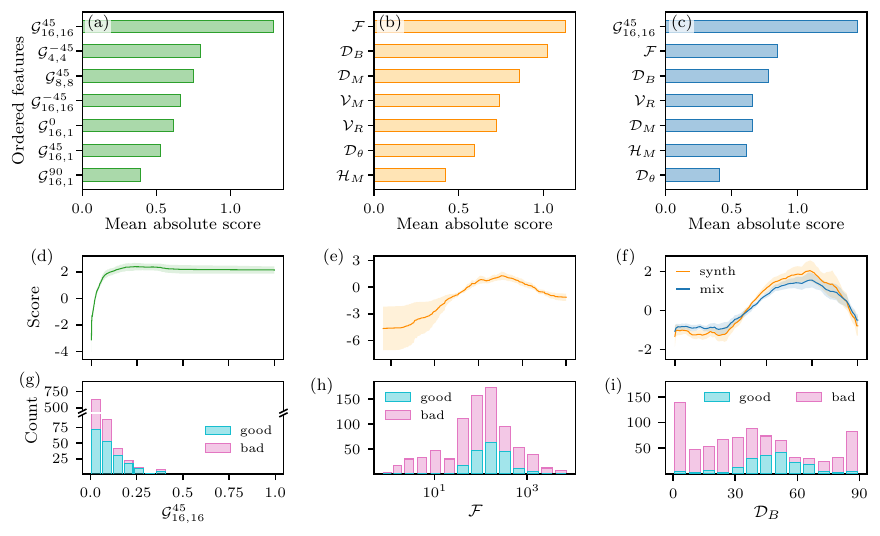}
  \caption{The feature importance plot for 
  (a) the Gabor filterbank vectorization approach,
  (b) the synthetic data modeling vectorization approach, and
  (c) the hybrid vectorization approach.
  The mean absolute score refers to the extent to which a feature contributes to all predictions.
  Feature curve for (d) the oriented Gabor feature $\mathcal{G}_{16,16}^{45}$, (e) the fit fitness feature $\mathcal{F}$ of the synthetic triangle fit, and (f) the diagonal boundary component $\mathcal{D}_{B}$ of the synthetic triangle fit for the synthetic data modeling approach shown in orange and for the hybrid approach shown in blue.
  The $x$-axis indicates the value attained by the feature [normalized to $(0,1)$ in (d) and (e)], and the score indicates the log odds towards good (positive) and bad (negative) classes for each respective feature.
  The histograms in (g-i) depict the relative class distribution as a function of the normalized feature value, with cyan representing good triangles and pink representing bad triangles.}
  \label{fig:global_exp}
\end{figure}

\subsection{Prediction interpretability}
\label{ssec:pre_interp}
To understand the EBM model decisions, we rely on the EBM's feature importance, as well as the plots of individual features.
The feature importance for all three cases we consider is shown in figure~\ref{fig:global_exp}(a)-(c).
The mean absolute score metric indicates the extent of the overall feature contribution to the model across all predictions for each feature in the model.
We also consider the feature curves, depicted in figure~\ref{fig:global_exp}(d)-(f), which show the log odds of being in a good vs. bad class, respectively, as a function of the feature values [normalized to $(0,1)$ for plots (d) and (e)].
The accompanying histograms, shown in figure~\ref{fig:global_exp}(g-i), provide the relative class distribution as a function of feature values.

Examining the overall feature importances for the Gabor method, shown in figure~\ref{fig:global_exp}(a), we see that the four dominant roles are occupied by oriented features $\mathcal{G}_{\sigma_x,\sigma_y}^\theta$ with $\theta = 45$.
The most important is a broad scale filter  $\mathcal{G}_{16,16}^{45}$, one of the filters characterizing the triangle region of the plots.

Examining the feature curve for the dominant filter, shown in figure~\ref{fig:global_exp}(d), we see that the model associates a strong $45^\circ$ response with good triangles.
The curve also reveals a threshold in the filter response below which a triangle is expected to be bad.
The histogram in figure~\ref{fig:global_exp}(d), showing the distribution of the good and bad features, further supports the discriminative nature of this filter, with $90.4~\%$ of bad data falling in the first bin.
It is notable that despite the higher performance of the Gabor-based technique, each $\mathcal{G}_{\sigma_x,\sigma_y}^\theta$ filter reveals only simple information about the extent of the presence of $45^\circ$ information. 
Thus, despite being highly discriminative, the information obtained from this family of filters does not directly provide information on how to adjust the experimental setup.

The feature importance plot for the synthetic data modeling approach, shown in figure~\ref{fig:global_exp}(b), reveals that this model also prioritizes features characterizing the triangle region, with $\mathcal{D}_{B}$ (the diagonal boundary component) and $\mathcal{D}_{M}$ (the diagonal magnitude component) being among the top three filters.
The most important feature here is the fit fitness $\mathcal{F}$, quantifying how close the vectorized representation matches the experimental data.
Looking at the feature curve for $\mathcal{F}$ depicted in figure~\ref{fig:global_exp}(e), we see that for a modest cost associated with fitting, that is when $10^2\lessapprox \mathcal{F}\lessapprox 10^3$, the log odds of belonging to the good class are positive. 
For bad triangles, the associated cost is either disproportionately low or high.
On the low end of the spectrum, this is likely due to the problem of fitting a bad triangle with only two rectangular regions being a significantly simpler optimization problem.
On the high end, the exceptionally high cost is generally associated with poor quality of the data.

\begin{figure}[t]
  \centering
  \includegraphics[width=0.95\linewidth]{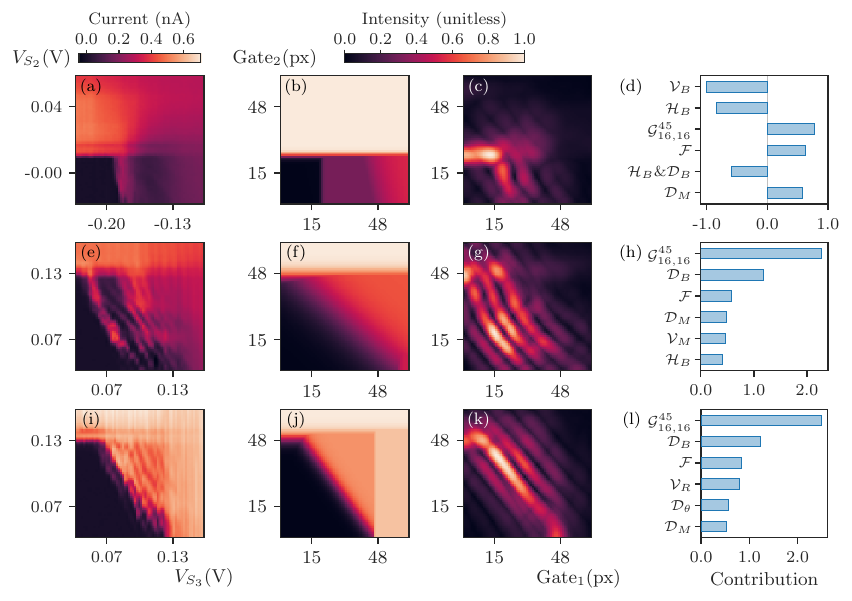}
  \caption{(a) A sample experimentally acquired bad scan.
  (e, i) Two sample experimentally acquired good scans.
  (b, f, j) The fitted synthetic triangle plots for scans (a), (e), and (i), respectively.
  (c, g, k) Gabor-filter-based local interpretation of scans is shown in (a), (e), and (i), respectively.
  (d, h, l) The feature importance for the hybrid vectorization method for scans (a), (e), and (i), respectively.
  The intercept coefficient $\alpha=4$ is removed from plots (d), (h), and (l) for clarity.}
  \label{fig:local_exp}
\end{figure}

The second most important feature in the synthetic data modeling approach is $\mathcal{D}_{B}$.
The $\mathcal{D}_{B}$ feature curve, shown in orange in figure~\ref{fig:local_exp}(f), indicates that the log odds that a triangle region belongs to a good plot increase with $\mathcal{D}_{B}$ once it extends past 28.
The log odds of a triangle region belonging to a good plot are positive when $35<\mathcal{D}_{B}<85$.
The $\mathcal{D}_{B}$ feature quantifies the extent to which the triangle region stretches in the screening gate-screening gate plot, as depicted in figure~\ref{fig:methods_vis}(c).
When $\mathcal{D}_{B}<35$, it is likely that the fitted triangle region overlaps with the horizontal and/or vertical sigmoid, and thus $\mathcal{D}_{B}$ would not be a reliable discriminant between good and bad class.
While theoretically vector $\mathcal{D}_{B}$ can extend to the bottom left corner in figure~\ref{fig:methods_vis}(c), for large values of $\mathcal{D}_{B}$ the hypotenuse becomes relatively short which makes finding a reliable fit challenging.
This dependency is confirmed by the feature plot, indicating that for $\mathcal{D}_{B}>85$, the log odds of a triangle region belonging to a good class fall below $0$.
The importance of the $\mathcal{D}_{B}$ filter is further evidenced by its dominant role in the hybrid approach, where it occupies the third most important position, see figure~\ref{fig:local_exp}(c). 

It is important to emphasize that, due to the way EBMs are trained, the feature curves are not the same as the marginal distribution of feature values. 
Rather, each curve is affected by the complete vector of features $\textbf{v}(\mathcal{I}_{exp})$ used to train the model.
This means that individual features cannot be treated as entirely independent from one another.
This is confirmed by the slight difference between the shape of the $\mathcal{D}_{B}$ curve for the synthetic data modeling approach vs. the hybrid approach, shown in figure~\ref{fig:global_exp}(f) in orange and blue, respectively.

The hybrid model achieves comparable performance to the Gabor-based model while including only a single Gabor feature in addition to the synthetic features.
It thus retains the improved intelligibility of the synthetic features without a significant reduction in accuracy. 

EBMs also give us the ability to examine predictions for individual images.
Figure~\ref{fig:local_exp}(a), (b), and (c) show an example of a bad image, its synthetic fit, and Gabor filter representation, respectively.
The synthetic triangle fit effectively captures the structural content of the experimental image, exhibiting no triangle region. 
The relative feature importance plot, shown in figure~\ref{fig:local_exp}(d), also agrees with this observation, indicating $\mathcal{V}_{B}$ (the vertical boundary component) and $\mathcal{H}_{B}$ (the horizontal boundary component) as the most dominant features contributing to the negative (i.e., bad class) prediction.  

Similarly, for good triangle regions, shown in figure~\ref{fig:local_exp}(e) and (i), the synthetic triangle fit also correctly captures the structural content, as depicted in figure~\ref{fig:local_exp} (f) and (j), respectively. 
In the relative feature importance plots, shown in figure~\ref{fig:local_exp}(h) and (i), respectively, the top three dominant features include the Gabor filter $\mathcal{G}_{16,16}^{45}$, $\mathcal{D}_{B}$, and $\mathcal{F}$, which is consistent with the global prediction for the good images class, albeit the 
order of the latter two features is swapped compared to figure~\ref{fig:global_exp}(c).
As expected, all three features strongly contribute to the positive class prediction. 

In all three examples, there is a direct observable relationship between the fit parameters and the visual features of the synthetic triangle.
In practical applications, these visual features can lead to thresholding decisions and derived quantities for choosing operating points in the voltage space. 
For example, in the case of figure~\ref{fig:local_exp}(i), a good operating point might be right inside the triangle region above the diagonal boundary, bisecting the area of the triangle at around $(V_{S_3}, V_{S_2})=(0.09~\si{\volt},0.09~\si{\volt})$ [at point (25,25) in figure~\ref{fig:local_exp}(j)]. 
Using the $\mathcal{D}_{B}$ and $\mathcal{D}_{\theta}$ with the knowledge of the boundaries of the image in voltage space, along with the $\mathcal{V}_{B}$ and $\mathcal{H}_{B}$, we can create a normal vector describing the direction in which the triangle is expanding. 
This normal vector describes the relative strengths of the associated screening gates. 
Ideally, the normal vector of the triangle would point symmetrically towards the bottom left corner, meaning that both screening gates have equal action on the forming current channel. 
But, more often than not, this vector tends to vary. 
By re-scaling the normal vector and defining the magnitude to begin at ($\mathcal{V}_{B}$,$\mathcal{H}_{B}$) with the end pointing to the diagonal boundary of the triangle, the vector can now encode size information. 
If we then reduce the magnitude by $10~\%$, we have an initial guess, inside the bounds of the triangle, for the operating point.
We can also use the vector to calculate the area of the triangle in voltage space, which can be thresholded to ensure it is large enough to proceed. 

If, on the other hand, the triangle plot looks like in figure~\ref{fig:local_exp}(a), and no triangle is predicted, we can enhance the chance of detecting one by increasing the voltage of gates over the current channel and taking the triangle plot once more. 
Hence, there are many derivable quantities from this model that could be used iterably to function as feedback in a larger tune-up procedure beyond just determining triangle existence.

\section{Conclusion and outlook}
\label{sec:conclusion}
In this work, we demonstrate an alternative approach to vectorizing image data for use with EBMs that relies on generating synthetic data that best fits the experimentally acquired scans.
We show that the new method is better adapted to the complexity of the triangle plots dataset.
While the original vectorization method, invoking the Gabor filterbank features, produces a model that performs well with the classification task, its interpretability and usefulness are limited as the resulting model does not comprehend the underlying structure of the data.
The alternative method, relying on fitting synthetic triangle plots to the experimental data, retains comparable accuracy, failing in about $2~\%$ more cases than the Gabor filterbank approach.
However, unlike the Gabor approach, it produces features that are strongly aligned with heuristics-based intuition about the data, resulting in qualitatively superior explanations.
As such, the features can be tied directly back to the scientific problem and used to adjust the experimental system to produce triangle plots of desirable quality. 
Future work might include integrating this analysis directly into an automated real-time QD tuning system. 

It is reasonable to assume that the Gabor technique excels due to its ability to represent the ridged pattern in the triangle region which is considered an important characteristic of a good scan.  
We see that including a single Gabor filter—capturing the presence and prevalence of a diagonal activity—allows us to recover nearly all of the performance without compromising the model's interpretability.
Alternatively, to avoid calculating the Gabor filterbank, this behavior could also be encapsulated within the synthetic approach by representing this pattern with a sinusoidal term in the synthetic triangle model.

The one disadvantage of the synthetic data modeling approach, when weighed against the Gabor filterbank, approach is the greater computational cost of the former. 
This is primarily due to the reliance on a global optimizer, whereas the Gabor-based approach only requires a small number of fixed filters. 
To remedy this, alternative similarity functions that would improve convergence and be less reliant on a global optimizer could be explored.
However, care must be taken to ensure that the qualitative fitness is not reduced. 

Finally, while all data used in this work comes from a quad-QD device with an overlapping gate architecture, we expect these results to be generalizable to an entire subclass of devices in Si/$\text{Si}_{x}\text{Ge}_{1-x}$ with different gate designs~\cite{Neyens2024, Zajac16-SGA, Philips22-UCS}.
Likewise, other electronic materials that can host gate-defined QDs, such as bi-layer graphene~\cite{Eich2018} and silicon-MOS structures~\cite{Stuyck2024}, can benefit from ML-enabled characterization and tuning as proposed in this work. 
The only requirement is that the tuning procedure must involve confining electrons to a 1D channel with a pair (or pairs) of gates so that other additional gates can then divide that channel into a chain of QDs.

\section*{Data availability statement}
\label{sec:data_availability}
\vspace{-5pt}
The data that support the findings of this study are openly available at the following URL/DOI: \href{https://doi. org/10.5281/zenodo.14549897}{https://doi. org/10.5281/zenodo.14549897}. 
Figure source files are openly available at the following URL/DOI: \href{https://doi. org/10.5281/zenodo.14589568}{https://doi. org/10.5281/zenodo.14589568}.

\section*{Acknowledgments}\label{sec:Acknowledgment}
\vspace{-5pt}
J. P. Dodson present address: HRL Laboratories, LLC, 3011 Malibu Canyon Road, Malibu, CA 90265, USA.
J. Corrigan present address: Intel Corp., Hillsboro, OR 97124, USA.
We acknowledge Patrick Walsh and Emily Joseph for experimental assistance. 
We acknowledge HRL Laboratories, LLC for support and L.F. Edge for providing one of the Si/Si$_{x}$Ge$_{1-x}$ heterostructures used in this work. 
JC acknowledges support from the National Science Foundation Graduate Research Fellowship Program under Grant No.~DGE-1747503 and the Graduate School and the Office of the Vice Chancellor for Research and Graduate Education at the University of Wisconsin-Madison with funding from the Wisconsin Alumni Research Foundation.
This research was sponsored in part by the Army Research Office (ARO) under Grant Nos.~W911NF-23-1-0110 and W911NF-17-1-0274.  
We acknowledge the use of facilities supported by NSF through the UW-Madison MRSEC (DMR-2309000). The views and conclusions contained in this paper are those of the authors and should not be interpreted as representing the official policies, either expressed or implied, of the U.S. Government or the ARO. 
The U.S. Government is authorized to reproduce and distribute reprints for Government purposes, notwithstanding any copyright noted herein. 
Any mention of equipment, instruments, software, or materials; it does not imply recommendation or endorsement by the National Institute of Standards and Technology.

\renewcommand{\thesection}{A}
\renewcommand{\thefigure}{A\arabic{figure}}
\setcounter{figure}{0}
\renewcommand{\thetable}{A\arabic{table}}
\setcounter{table}{0}
\renewcommand{\arraystretch}{1}
\section*{Appendix}\label{sec:appendix}
\subsection{Misclassified data: A post-hoc analysis}
\label{app:misclass} 
\begin{figure}[t]
  \centering
  \includegraphics[width=0.95\linewidth]{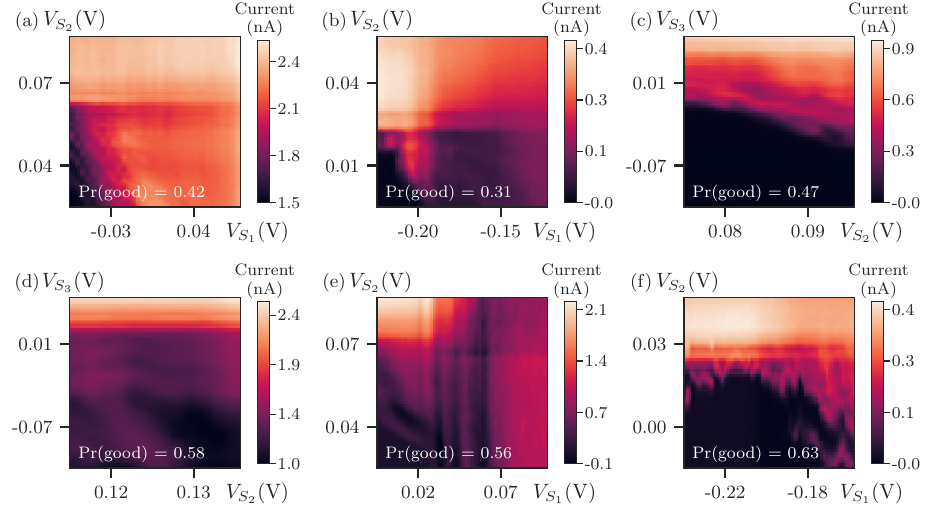}
  \caption{Sample ambiguous triangle plots.
  (a)--(c) Good-class triangle plots classified by the model as bad.
  (d)--(e) Bad-class triangle plots classified by the model as good.}
  \label{fig:misclass}
\end{figure}

\vspace{-5pt}
While the synthetic triangles provide a good representation of the crude features captured in experimental data for the majority of the data, there are cases where the vectorized representation is misclassified by the EBM.
There are two types of misclassifications: classifying a triangle plot with a well-pronounced triangle region as bad [false-negative, see figure~\ref{fig:misclass} panels (a)--(c)] and classifying a plot without well-defined triangle region as good [false-positive, see figure~\ref{fig:misclass} panels (d)--(e)].

There are several reasons for such misclassifications. 
In figure~\ref{fig:misclass}(a) and (b), the plots are excessively accumulated, which would usually indicate a good sample if the user were to zoom out, and there is not sufficient diagonal activity for the sample to be valid. 
In figure~\ref{fig:misclass}(a) the scan window seems too small for the size of the triangle plot. 
In figure~\ref{fig:misclass}(b), on the other hand, we see the shadow of a second channel influencing the readout. 
This is a rare breakdown of the model's assumptions where the channels of the QD device were assumed to be independent. 
The ohmic potential is typically proportional to the maximum current in the image.
With high accumulation, the ohmic potential can drift dynamically depending on multiple channels simultaneously since the measured current takes the path of the lowest resistance. 
This issue only plagues the upper channel of this device with charge sensors, see figure~\ref{fig:dev_data}(a), since they are not separated by a gate-like isolation of the quad-QD channel on the bottom of the device.
To resolve this with the same model, the experimental setup would need to know exactly when the ohmic potential started drifting and compensate for both halves of the charge sensor to keep exactly half the current flowing out through both channels equally.
In figure~\ref{fig:misclass}(c), the slope indicating the formation of coulomb blockade is relatively weak but technically visible, which, to an experimentalist, indicates that the triangle plot is good. 
However, in all three plots, one of the channels seems to be missing, which contributed to them being classified as bad even though the triangle region is present in all of them. 

In figure~\ref{fig:misclass}(d), there is a substantial region with no current inside the triangle region, which seems to suggest that the channel is not accumulated correctly. 
In figure~\ref{fig:misclass}(e), the device is tuned to the wrong regime for the triangle plot where the blockade is controlled by one screening gate more than the other. 
This can be seen by the fringes being vertical rather than diagonal.
Because plots in figure~\ref{fig:misclass}(d) and (e) have erroneous dropout regions, they should be considered bad. 
However, because the presence of the significant diagonal behavior has been vectorized through the synthetic triangle modeling, the EBM classified these images as good. 
In the plot depicted in figure~\ref{fig:misclass}(f), the strong diagonal and ridged pattern indicates a possibly shocked device.
Yet, the diagonal rigid region is incorrectly vectorized as a proper triangle, which again leads to incorrect classification by EBM.

\section*{ORCID iDs}
\label{sec:orcids}
Daniel Schug~\orcidlink{0009-0001-3758-501X} \href{https://orcid.org/0009-0001-3758-501X}{https://orcid.org/0009-0001-3758-501X} \\ 
Tyler J Kovach~\orcidlink{0009-0007-0807-7300} \href{https://orcid.org/0009-0007-0807-7300}{https://orcid.org/0009-0007-0807-7300} \\ 
Jared Benson~\orcidlink{0009-0009-1673-5259} \href{https://orcid.org/0009-0009-1673-5259}{https://orcid.org/0009-0009-1673-5259} \\ 
J P Dodson~\orcidlink{0000-0003-4265-5024} \href{https://orcid.org/0000-0003-4265-5024}{https://orcid.org/0000-0003-4265-5024} \\
M A Eriksson~\orcidlink{0000-0002-3130-9735} \href{https://orcid.org/0000-0002-3130-9735}{https://orcid.org/0000-0002-3130-9735} \\
Justyna P Zwolak~\orcidlink{0000-0002-2286-3208} \href{https://orcid.org/0000-0002-2286-3208}{https://orcid.org/0000-0002-2286-3208}

\newcommand{\newblock}{}

\end{document}